\documentclass{article}

\usepackage{microtype}
\usepackage{graphicx}
\usepackage{subfigure}
\usepackage{booktabs} 

\usepackage{amsmath}
\usepackage{amssymb}
\usepackage{dsfont}
\usepackage{mathtools}
\usepackage{multirow}
\usepackage{float,paralist}

\usepackage{hyperref}

\usepackage[accepted]{icml2021}

\newcounter{remark}
\newenvironment{Remark}[1][]{\refstepcounter{remark}\par\medskip
   \noindent \textit{Remark~\theremark. #1} \rmfamily}{\medskip}
\mathtoolsset{showonlyrefs}

\def\vone{\textbf{1}}
\def\R{\mathbb{R}}
\def\E{\mathbb{E}}
\DeclarePairedDelimiter\abs{\lvert}{\rvert}

\icmltitlerunning{You Only Sample (Almost) Once: Linear Cost Self-Attention Via Bernoulli Sampling}

\begin{document}

\twocolumn[
\icmltitle{You Only Sample (Almost) Once: Linear Cost Self-Attention Via Bernoulli Sampling}



\icmlsetsymbol{equal}{*}

\begin{icmlauthorlist}
\icmlauthor{Zhanpeng Zeng}{wisc}
\icmlauthor{Yunyang Xiong}{wisc}
\icmlauthor{Sathya N. Ravi}{uic}
\icmlauthor{Shailesh Acharya}{amfam}
\icmlauthor{Glenn Fung}{amfam}
\icmlauthor{Vikas Singh}{wisc}
\end{icmlauthorlist}

\icmlaffiliation{wisc}{University of Wisconsin, Madison, USA}
\icmlaffiliation{uic}{University of Illinois, Chicago, USA}
\icmlaffiliation{amfam}{American Family Insurance, Madison, USA}

\icmlcorrespondingauthor{Zhanpeng Zeng}{zzeng38@wisc.edu}

\icmlkeywords{Machine Learning, ICML}

\vskip 0.3in
]



\printAffiliationsAndNotice{}  

\begin{abstract}
Transformer-based models are widely used in natural language processing (NLP). Central to the transformer model is the self-attention mechanism, which captures the interactions of token pairs in the input sequences and depends quadratically on the sequence length. Training such models on longer sequences is expensive. In this paper, we show that a Bernoulli sampling attention mechanism based on Locality Sensitive Hashing (LSH), decreases the quadratic complexity of such models to linear. We bypass the quadratic cost by considering self-attention as a sum of individual tokens associated with Bernoulli random variables that can, in principle, be sampled at once by a single hash (although in practice, this number may be a small constant). This leads to an efficient sampling scheme to estimate self-attention which relies on specific modifications of LSH (to enable deployment on GPU architectures). We evaluate our algorithm on the GLUE benchmark with standard 512 sequence length where we see favorable performance relative to a standard pretrained Transformer. On the Long Range Arena (LRA) benchmark, for evaluating performance on long sequences, our method achieves results consistent with softmax self-attention but with sizable speed-ups and memory savings and often outperforms other efficient self-attention methods. Our code is available at \url{https://github.com/mlpen/YOSO}.
\end{abstract}

\section{Introduction}

The Transformer model \citep{vaswani2017attention} is incredibly effective across natural language processing (NLP) applications including machine translation \citep{vaswani2017attention}, language inference \citep{devlin2018bert} and paraphrasing \citep{raffel2020exploring}. 
Transformer-based models such as BERT \citep{devlin2018bert} are pretrained in an unsupervised manner and later finetuned on different downstream tasks, often providing state-of-the-art performance on standard benchmarks. While such models have strong empirical performance, their computational/memory requirements remain high. Consequently, in the NLP setting, many current models have certain constraints on the sequence length, e.g., BERT and other transformer-based language models \citep{yang2020xlnet, liu2019roberta} limit the sentence length to be at most $512$, although recent results have reported success with longer sequences based on interesting efficiency-focused strategies \citep{beltagy2020longformer,zaheer2020big,xiong2021nystromformer}.

Multi-Head Self-Attention is central to Transformer based models and provides a flexible global receptive field to exchange information among input tokens. While self-attention provides various benefits, it is also a bottleneck when training with long sequences. In particular, the output of self-attention is a combination of all tokens where coefficients are determined by the similarities among tokens. This is beneficial, but involves a sizable resource footprint. When the sequence length is $n$, one incurs a $O(n^2)$ complexity in both time and memory to compute pairwise similarities among all input tokens. This quadratic cost 
restricts its use in applications where capturing long term context dependencies is important,
and has motivated 
many ongoing efforts to mitigate the resource needs of such models. 

Our work, also seeks to address the aforementioned issues, and is inspired by ideas of importance sampling via hashing-based sampling strategies \citep{spring2017new, charikar2017hashing}. We propose a Bernoulli based sampling scheme to approximate self-attention, which scales linearly with the input sequence length. We view self-attention as a sum of individual tokens associated with Bernoulli random variables whose success probability is determined by the similarities among tokens. 
In principle, we can sample all Bernoulli random variables at once with a single hash.
It turns out that the resultant strategy (You Only Sample Almost Once, YOSO-Attention) is more amenable to an efficient/backpropagation friendly implementation, and exhibits a favorable performance profile in experiments.

\section{Related Works}

We first review a commonly used form for self-attention and then the most relevant ideas that also focus on efficiency aspects of Transformer models. Then, we summarize a
distinct line of work based on sampling which directly inspires our strategy for approximating softmax calculation.


\subsection{Self-Attention}

Self-attention is a scaled dot-product attention mechanism to capture token dependencies in the input sequence, which can be defined as,
\begin{equation}
\begin{split}
\mathcal{A}(Q, K, V) &= \text{softmax}\left( \underbrace{\frac{(Q W_Q) (K W_K)^T}{\sqrt{d_h}}}_{\mathcal{P}} \right)V W_V \\
&= D_{\mathcal{P}} \exp\left( \mathcal{P} \right) V W_V
\end{split}
\end{equation}\label{eq:softmax-attention}
where $Q, K, V \in \R^{n \times d}$ are embedding matrices from the input sequence, and called  queries, key and values respectively. Here, $n$ is the input sequence length, $d$ is the embedding dimension of each token, $W_Q, W_K, W_V \in \R^{d \times d_h}$ are learned parameter matrices, $d_h$ is the head dimension, and $D_{\mathcal{P}}$ is a $n\times n$ diagonal matrix which normalizes each row of the $\exp{(\mathcal{P})}$ matrix such that the row entries sum up to $1$. For simplicity, we overload the notations for $Q, K, V$ to denote $Q W_Q, K W_K, V W_V$ in our description below.


\textbf{Multi-Head Self-Attention}.  Multi-Head self-attention in Transformers runs through the scaled dot-product attention multiple times and the attention outputs are concatenated to help the model capture information from multiple representation subspaces \cite{vaswani2017attention}. Multi-Head Self-attention can be formally written as, 
\begin{equation}
\text{MultiHead}(Q, K, V) = \sum_{a=1}^h \mathcal{A}_a(Q, K, V) W_a
\end{equation}\label{eq:multihead-attention}
where $h$ is the number of heads, $\mathcal{A}_a, a = 1, \dots, h$ are heads with different parameter matrices, and $W_a \in \mathbb{R}^{d_h \times d}$, $a = 1, \dots, h$ are learnable transformations. 

\textbf{Self-Attention Bottleneck}. A bottleneck in self-attention is calculating the softmax matrix, $\text{softmax}(\mathcal{P})$, which requires all pairwise input token similarities. 
%


\subsection{Efficient Transformers}

Recent proposals have identified a number of 
ways to reduce the quadratic cost of self attention. 
%
Linformer \cite{Wang2020LinformerSW} shows that using a low-rank assumption, self-attention can be approximated via random projections along the sequence length dimension. The authors replace random projections with learnable linear projections and achieve a $O(n)$ complexity via a fixed projection dimension. 
Linear Transformers \citep{Katharopoulos2020TransformersAR} replace the softmax activation by applying a separable activation on the queries and keys. 
Based on the connection between the softmax activation and the Gaussian kernel, Performer \citep{choromanski2020rethinking} and Random Feature Attention \citep{peng2021rfa} approximate  softmax as the dot product of finite dimensional random feature vectors with a guarantee of convergence (to softmax). 
Nystr\"{o}mformer \citep{xiong2021nystromformer}, on the other hand, uses a landmark-based Nystr\"{o}m method to approximate the attention matrices. 
These methods achieve $O(n)$ complexity by avoiding direct calculation of attention matrices. 
Multiple approaches have been developed to exploit  sparsity and structured patterns of attention matrices observed  empirically. This line of work includes Sparse Transformer \citep{Child2019GeneratingLS}, Longformer \citep{beltagy2020longformer}, and Big Bird \citep{zaheer2020big}, which involve time and memory complexity of $O(n \sqrt{n}), O(n), O(n)$ respectively. 
The Reformer method \citep{Kitaev2020ReformerTE}, which is related to our work, also utilizes the sparsity of self-attention. But instead of predetermining a sparsity pattern, it uses Locality Sensitive Hashing (LSH) as a tool to approximate nearest neighbor search, and dynamically determines the sparsity pattern to achieve $O(n \log(n))$ complexity. In contrast, our approach takes advantage of the connection of query-key similarity to the LSH collision probability which is used to directly estimate self-attention without relying on sparsity.

\subsection{Importance Sampling}


We focus on developing an efficient low variance estimator of self-attention. Since self-attention can be thought of as integrating tokens over a softmax distribution, 
to estimate this integral, one could use importance sampling, see \citep{press2007numerical}. 
Using importance sampling 
will allow drawing samples from a uniform distribution and avoids sampling from the softmax distribution directly (which is harder). But this leads to a high variance estimate since the softmax distribution is usually concentrated in a small region. 

\textbf{LSH-based Importance Sampling}. Consider the case when the angular distance between a key and a query is small. In this case, the similarity (between the key and the query) as well as the softmax probability will be large. 
When viewed through the lens of a nearest neighbor retrieval, the above property coincides with a large collision probability of high similarity key-query pairs, assuming that the neighbor retrieval is implemented via LSH. 
Motivated by the link between softmax probability $p$ and LSH collision probability $q$, \citet{spring2017new} and \citet{charikar2017hashing} suggest using LSH as an efficient sampler for low variance softmax estimators, which can be adopted for self-attention approximation.

\textbf{(a)}  \citet{spring2017new} propose approximating softmax by sampling a set, $S$, a collection of neighboring keys for each query formed by the union of colliding keys using $m$ hash tables. The estimator is computed using
${\abs{S}}^{-1} \sum_{j \in S} \frac{p(Q_i, K_j)}{q(Q_i, K_j)} V_j$,
where $Q_i$ is a query vector, $K_j, V_j$ are key and value vectors in the sampling set $S$, and $p(\cdot, \cdot)$ and $q(\cdot, \cdot)$ are softmax probability and collision probability of given pairs. This procedure involves importance sampling without replacement, which leads to a dependency among the samples. Deduplication (avoiding double counting) requires memory to store keys in each hash table and runtime to deduplicate keys for each query. If the size of hash buckets is skewed, the (GPU) memory needs depend on the size of the hash bucket and the runtime depends on the size of $S$. 

\textbf{(b)} \citet{charikar2017hashing} provide a Hash based Estimator to simulate a proposal distribution for importance sampling via LSH, which can be easily applied in the context of softmax. For each hash table, a key is uniformly selected from the bucket that the query is hashed to, for simulating a draw from a proposal distribution. The estimate is computed as 
${m}^{-1} \sum_{k = 1}^m \frac{p(Q_i, K_j)\abs{H_k(Q_i)}}{q(Q_i, K_j)} V_j$, 
where $\abs{H_k(Q_i)}$ denotes the size of hash bucket in the $k$-th hash table which $Q_i$ is hashed to. This simulates $m$ samples drawn with replacement from the proposal distribution. However, the probability of one key being sampled depends not only on (i) the angular distance to the query but also (ii) the number of keys within the hash bucket, leading to a sampling dependency among all keys. Further, using it for self-attention causes a dependence between the sparsity in the softmax matrix and the number of hashes used. Specifically, the number of tokens that each query can attend to is bounded by the number of hashes: the procedure samples at most one distinct key for each hash table and so, it adds one additional nonzero to the softmax matrix, at most. 

\textbf{LSH-based Importance Sampling: practical considerations}. While LSH-based importance sampling exploits the agreement between high probability $p(\cdot, \cdot)$ and high collision probability $q(\cdot, \cdot)$, the alignment is not perfect. Samples from the proposal distribution must be reweighted to compensate for the difference. Further, for different queries, the likelihood ratios between the softmax distribution and the proposal distribution w.r.t. a single key are different. Therefore, a reweighing has to be done {\em during} querying. Although maintaining hash tables for storing keys is not a major problem in general, the high memory cost for hash tables and computation time for reweighing noticeably influences efficiency when applied to self-attention. 
To {\bf summarize}, we find that a direct application of LSH-based importance sampling in the deep learning context may not lead to an efficient self-attention scheme.

\section{YOSO Attention}

\subsection{YOSO Attention}

While the softmax computation bottleneck can be alleviated through LSH-based importance sampling, these approaches are not very efficient on GPUs. Our LSH-based Bernoulli sampling offers benefits here. Instead of using LSH to simulate sampling from a proposal distribution over tokens, we view attention as a {\em sum} of tokens associated with Bernoulli random variables. This modification relates better with LSH and less with LSH-based importance sampling -- {\em the probability of one query colliding with a key is not based on other keys}. This strategy helps avoid the sampling dependency problem in LSH-based importance sampling and offers us an opportunity to develop a strategy more amenable to GPUs.

\vskip -0.1in
\begin{Remark}
We assume that the input keys and queries of self-attention are unit length -- to allow treating dot-product similarity in self-attention and  cosine similarity in LSH similarly. This is simple  using \citet{neyshabur2015symmetric}: a variable $\tau$ is used to bound the squared $\ell_2$ norm of all queries and keys and to reconstruct new unit length keys and queries while preserving their pairwise similarities. Then, we can work with the softmax matrix in angular distance metric and derive our algorithm.
\vskip -0.1in
\end{Remark}

 

\begin{figure*}[h]
\centering
\vspace{-0.05in}
\includegraphics[width=0.9\textwidth]{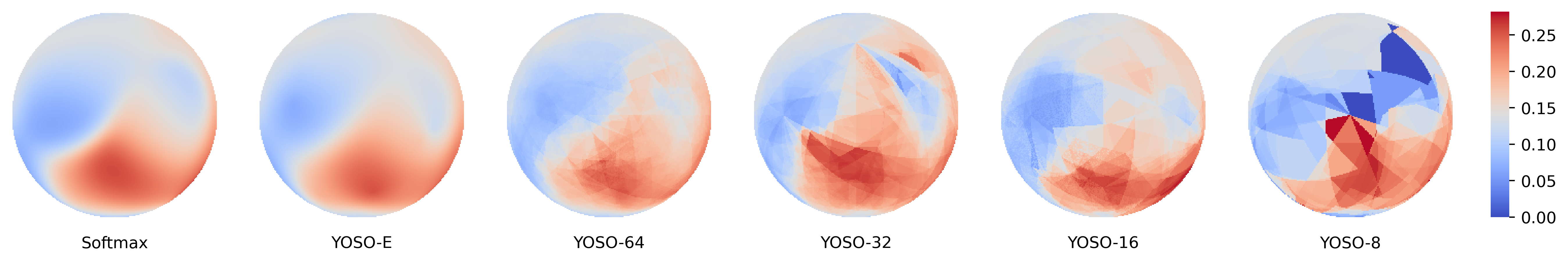}
\vspace{-0.1in}
\caption{A high-level overview of YOSO-Attention. YOSO-m denotes YOSO-Attention with $m$ samples (hashes) and YOSO-E denotes the expectation of YOSO over hash functions. Softmax denotes the softmax self-attention. We visualize YOSO-m, YOSO-E, and Softmax by randomly generating $K \in \R^{32 \times 3}, V \in \R^{32 \times 1}$ and using all unit vectors $Q_i \in S^2$ to compute their output values on the surface of 3-dimensional sphere to demonstrate how YOSO-m approximates YOSO-E and the high similarity between YOSO-E and Softmax. 
}
\label{fig:func-approx}
\end{figure*}

\begin{figure}[!b]
\centering
\vspace{-0.05in}
\includegraphics[height=1.88in]{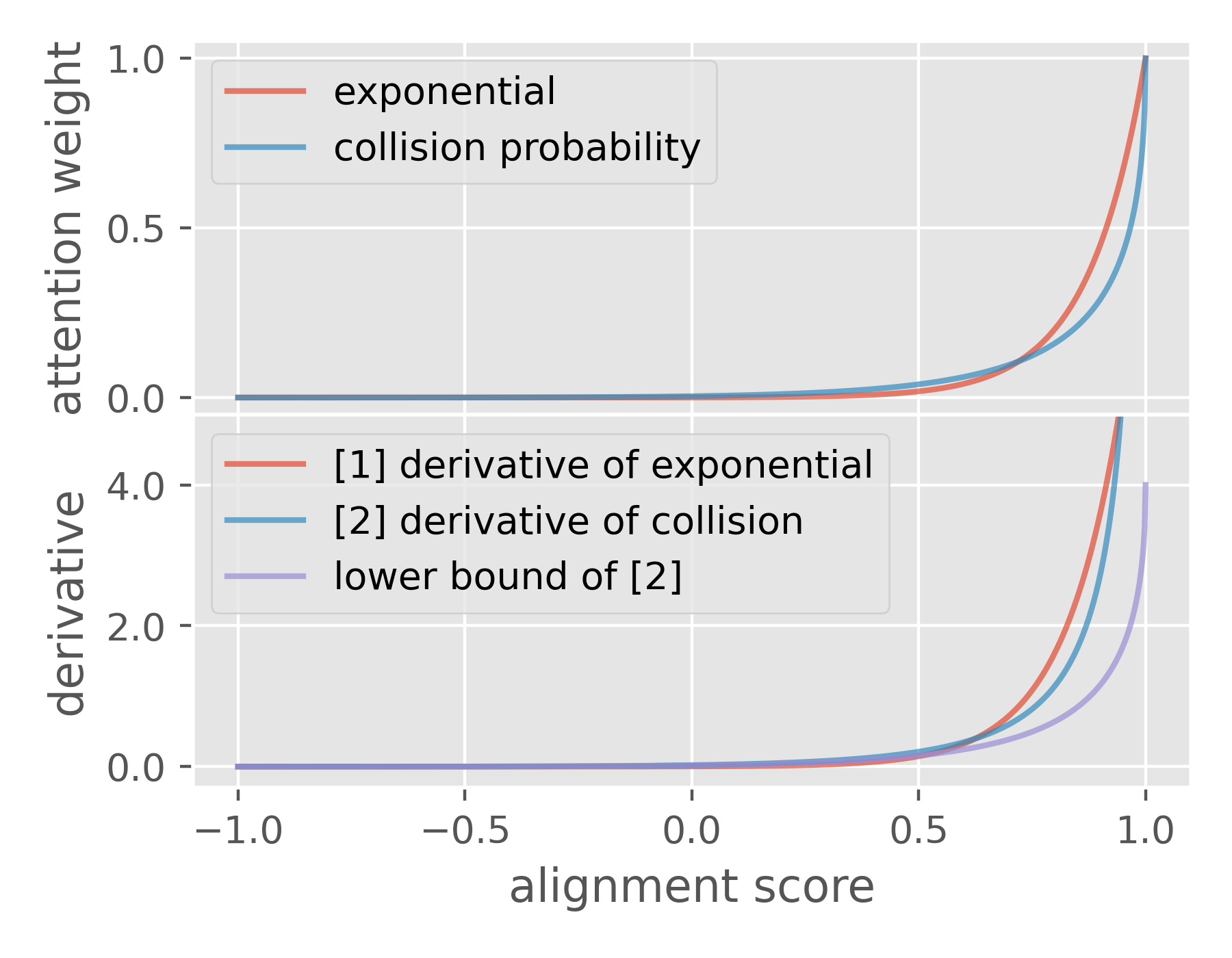}
\vspace{-12pt}
\caption{We compare attention weights using $\exp(\tau (x - 1))$ with the collision probability of concatenating $\tau$ hyperplane hashes \citep{charikar2002similarity} $(1 - \arccos(x) / \pi)^{\tau}$ for $\tau = 8$. We plot $\exp(\tau (x - 1))$ so that the range is between $0$ and $1$ but without changing the actual attention weights in softmax. We also plot the derivative of exponential function and of collision probability, as well as a lower bound we will use later during backpropagation. }
\label{fig:comparison-exp-collision}
\end{figure}

\textbf{Self-Attention via Bernoulli Sampling}. 
We aim to approximate self-attention, which uses a softmax matrix to capture the context dependency among tokens via their pairwise similarities. Assuming that we can represent this context dependency {\em directly} using collision probability $q(\cdot, \cdot)$, no reweighting is required if the proposal and target distributions are the same. This means that challenges discussed in LSH-based importance sampling do not exist. So the coincidence of softmax probability $p(\cdot, \cdot)$ and LSH collision probability $q(\cdot, \cdot)$ makes $q(\cdot, \cdot)$ a sensible starting point for approximating self-attention. 
Specifically, to model dependency based on similarity, the collision probability aligns well with the exponential function in softmax in the domain of interest $[-1, 1]$ in Figure \ref{fig:comparison-exp-collision}: both functions have positive zeroth, first and second order derivatives. 

Note that (a) positive zeroth order derivative indicates that the dependency is positive, (b) positive first order derivative ensures that the dependency based on similarity is monotonic, and (c) positive second order derivative means that 
the attention weight will rapidly increase and dominate others as the similarity increases.
This leads us to hypothesize that a collision-based self-attention may be as effective as softmax-based self-attention. 
It can be formulated as, 
\begin{equation}\label{eq:sum-rv}
\sum_{j = 1}^n \mathcal{B}(Q, K)_{i,j} V_j
\end{equation}
where $\mathcal{B}(Q, K)_{i,j}$ is a Bernoulli random variable where the success probability is given by the collision probability of $Q_i$ with the key $K_j$. Hence, it can be determined by the similarity between $Q_i, K_j$. 

In a single hash, each $\mathcal{B}(Q, K)_{i,j}$ generates a realization to determine whether the corresponding token will be part of attention output or not. Conceptually, when sampling from the softmax distribution, only one token is sampled as the attention output. In contrast, Bernoulli sampling determines whether each individual token is a part of the attention output. In principle, to determine the context dependency among tokens, you only need to sample once (YOSO) using a single hash to generate {\em realizations of all Bernoulli random variables}, $\mathcal{B}(Q, K)_{i,j}, i, j = 1, \dots, n$. Specifically, when keys are hashed to a hash table using a single hash, the realization of $\mathcal{B}(Q, K)_{i,j}$ for each query $Q_i$ will be $1$ if $Q_i$ collides with $K_j$, else it is $0$.
To our knowledge, using LSH collision probability to replace softmax dependencies for self-attention 
in this way has not been described before. 

\textbf{YOSO-Attention}. 
By replacing softmax dependency with Bernoulli random variables and using LSH as an efficient sampler to estimate the success probability, we obtain an efficient self-attention (YOSO-Attention) to approximate softmax-based self-attention. 
\begin{equation}
\text{YOSO}(Q, K, V) = \mathcal{B}(Q, K) V
\label{eq:yoso-formulation}
\end{equation}
where $\mathcal{B}(Q, K)$ is the Bernoulli random matrix. 
\begin{equation}
\mathcal{B}(Q, K)_{i,j} = \mathds{1}_{f(Q_i) = f(K_j)}
\end{equation}
where $f$ is a hash function. The expectation of $\mathcal{B}(Q, K)_{i,j}$ is 
\begin{equation}
\mathbb{E}[\mathcal{B}(Q, K)_{i,j}] = \left(1 - \frac{\arccos(Q_i K_j^T)}{\pi}\right)^{\tau}
\label{eq:yoso-expectation}
\end{equation}
The variance of a Bernoulli random variable is simply: 
\begin{equation}
\text{var}[\mathcal{B}(Q, K)_{i,j}] = \mathbb{E}[\mathcal{B}(Q, K)_{i,j}] (1 - \mathbb{E}[\mathcal{B}(Q, K)_{i,j}])
\label{eq:yoso-variance}
\end{equation}
While a single sample would work in estimating attention output, in practice, the actual output of YOSO-Attention can be the average of output from $m$ samples to lower the estimation variance, where $m$ is a small constant. The high-level overview of our method is demonstrated in Figure \ref{fig:func-approx}. For LSH, each sample (hash) is a space partitioning of the input space. The $V_j$'s associated with $K_j$'s in the same partition are summed together. The partitions give a coarse representation of $\E[\text{YOSO}(\cdot, K, V)]$. As $m$ increases, the average of $m$ representations converges to $\E[\text{YOSO}(\cdot, K, V)]$.

\vskip -0.1in
\begin{Remark}
Our proposed method enjoys multiple advantages explicitly noted recently in Performer \citep{choromanski2020rethinking}, which is desired for self-attention: \textbf{(a)} The attention weights are always positive, which make it a robust self-attention mechanism. In YOSO, the attention weights are always in $[0, 1]$, which means that it  is also numerically stable. \textbf{(b)} The variance goes to zero as attention weight approaches zero. In YOSO, the variance of attention weights are always upper bounded by the attention weights themselves, making the approximation error easily controllable. 
\vskip -0.1in
\end{Remark}

\begin{figure*}[!htbp]
\centering
\vspace{-0.05in}
\includegraphics[width=0.925\textwidth]{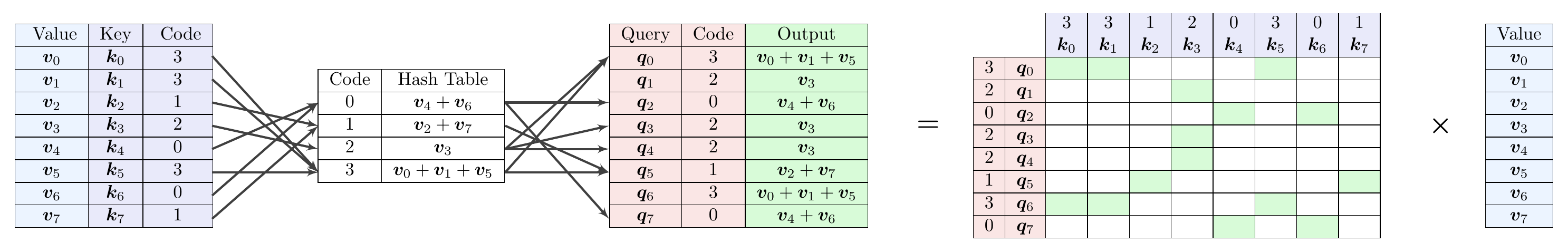}
\vspace{-0.1in}
\caption{Overview of YOSO-Attention algorithm. The hash table stores the sum of values associated with hashed keys. }
\label{fig:yoso-algorithm}
\end{figure*}

\textbf{Normalizing Attention}. In softmax self-attention, each row of the softmax matrix is normalized so that the dependencies sum up to $1$. We discussed above how the pairwise query-key dependencies can be estimated using Bernoulli sampling. We now describe how to normalize the dependency in our method as softmax self-attention. 
We can first estimate the dependencies and then normalize them using the sum of estimated dependencies estimated by $\mathcal{B}(Q, K) \vone$ where $\vone$ is a vector of all entries being $1$. $\mathcal{B}(Q, K) \vone$ can be computed by \eqref{eq:yoso-formulation} by plugging $\vone$ into $V$. To make the estimation of self-attention more efficient, we adopt a $\ell_2$ normalization on the attention output, similar to use of $\ell_2$ normalization for word embedding in 
\citet{levy-etal-2015-improving}. 
Thus, attention outputs are invariant to  scaling, $\mathcal{B}(Q, K) \vone$, under $\ell_2$ normalization. Therefore, we have, 
\begin{equation}
\text{N-YOSO}(Q, K, V) = \ell_2(\mathcal{B}(Q, K) V)
\end{equation}\label{eq:normalized-yoso}
Empirically, as expected, we find that the $\ell_2$ normalization does not affect the performance of our method (discussed in the experiments). 

\subsection{LSH-based Bernoulli Sampling}

Now, we discuss how to actually implement the idea of using Bernoulli sampling to approximate self-attention. While a standard LSH procedure can be used, maintaining hash tables to store keys is inefficient on a GPU -- the GPU memory size required for hash table cannot be predetermined and the workload might be skewed due to skewed bucket sizes. 
Due to how our Bernoulli sampling is set up, it turns out that simply saving the summation of values corresponding to hashed keys is sufficient (instead of storing a full collection of hashed keys). 

\textbf{Overview}. An outline of our algorithm is shown in Figure \ref{fig:yoso-algorithm}. To compute $Y = \mathcal{B}(Q, K) V$, the procedure proceeds as follows. We sample a hash function $f$ and create a hash table $H \in \R^{2^{\tau} \times d}$ representing $2^\tau$ $d$-dimensional buckets. For each key $K_j$, we add the value $V_j$ to the bucket whose index is the hash code $f(K_j)$, denoted as $H_{f(K_j)}$, 
\begin{equation}
H_{f(K_j)} \gets H_{f(K_j)} + V_j
\quad \text{}
\end{equation}

Note that the size of $H$ is $O(2^{\tau} d)$ and is {\bf independent} of which bucket keys are hashed. With all keys processed, for each query $Q_i$, we maintain an output vector $Y_i$ initialized to $0$. Then, we allocate the bucket in $H$ using $f(Q_i)$ and use $H_{f(Q_i)}$ as the attention output $Y_i$ for $Q_i$. 
Therefore, each final output $Y_i$ can be computed as, 
\begin{equation}
Y_i = \sum_{j = 1}^n \mathds{1}_{f(Q_i) = f(K_j)} V_j = \sum_{j = 1}^n \mathcal{B}(Q, K)_{i, j} V_j
\end{equation}

\vspace{-10pt}
\begin{Remark}
The memory and time complexity of this algorithm are $O(m 2^{\tau} d)$ and $O(n m d)$ respectively, In addition, both time and memory are independent of the size of hash buckets. We can further improve the memory complexity to $O(m 2^{\tau})$ by reusing the hash table and processing a few dimensions each time without increasing the time complexity. The constant $\tau$ is a hyperparameter that controls the decay rate of attention weights with respect to the angular distance between query and key. 
\end{Remark}
\vspace{-5pt}

\textbf{Speed-up}. While not essential, we find that a fast random projection for computing the LSH hash code is beneficial, since this step takes a large portion of the overall runtime. As suggested by \citet{alex2015practical}, we use the approximated random projection to reduce time complexity to $O(n m \tau \log_2(d))$, allowing fast computation of hash codes (details in the appendix).



\subsection{Backpropagation}

For training, we also need to show that backward propagation steps for YOSO-Attention are feasible. Here, we discuss this last component of YOSO-Attention which enables end-to-end  efficient training. 

\begin{table}[!b]
\centering
\begin{small}
\begin{tabular}{lcc}
\toprule
Time & Forward     & Backward \\
\midrule
Softmax  & $O(n^2 d)$  & $O(n^2 d)$     \\
YOSO     & $O(n m \tau \log_2(d) + n m d)$  &  $O(n m d^2)$  \\
\toprule
Memory   & Forward     & Backward  \\
\midrule
Softmax  &  $O(n^2)$   &     $O(n^2)$         \\
YOSO     &  $O(n m \tau + m 2^{\tau})$    &    $O(m 2^{\tau})$       \\
\bottomrule
\end{tabular}
\end{small}
\caption{Time/memory complexity of self-attention and YOSO-attention in forward/backward computation}
\label{tab:complexity}
\end{table}

For backpropagation, the gradient of the loss $L$ w.r.t. $V$ can be estimated similar to \eqref{eq:yoso-formulation},  
\begin{equation}
\begin{split}
\nabla_{V} L &= ((1 - \frac{\arccos(Q K^T)}{\pi})^{\tau})^T (\nabla_{\text{YOSO}} L) \\
&\approx \mathcal{B}(K, Q) (\nabla_{\text{YOSO}} L)
\end{split}
\label{eq:gradient-v}
\end{equation}

The gradients of $L$ w.r.t. $Q, K$ are similar, so we only provide the expression for $Q$, 
\begin{equation}
\begin{split}
\nabla_{Q} L =& [\left((\nabla_{\text{YOSO}} L) V^T \right) \odot \left(\tau (1 - \frac{\arccos(Q K^T)}{\pi})^{\tau - 1}\right)  \\
& \oslash \left(\pi \sqrt{1 - (Q K^T)^2}\right)] K 
\end{split}
\label{eq:gradient-q}
\end{equation}
where $\oslash, \odot$ are element-wise division and multiplication. One issue with the true gradient is that it goes to infinity as the alignment score between the query and the key approaches $1$, which might lead to divergence. 
To avoid this numerical issue, we use a lower bound of the actual derivative of the collision probability, 
\begin{equation}
\begin{split}
\hat{\nabla}_{Q} L &= [ \left((\nabla_{\text{YOSO}} L) V^T\right) \odot \frac{\tau}{2} (1 - \frac{\arccos(Q K^T)}{\pi})^{\tau} ] K \\
&\approx [ \left((\nabla_{\text{YOSO}} L) V^T\right) \odot \frac{\tau}{2} \mathcal{B}(K, Q) ] K
\end{split}
\label{eq:pseudo-gradient-q}
\end{equation}
The empirical behavior is shown in Figure \ref{fig:comparison-exp-collision}, and it can be efficiently estimated via a variation of LSH-based Bernoulli Sampling. Specifically, note that the approximation can be decomposed into sum of $d$ LSH-based Bernoulli Sampling,
\begin{equation}
(\hat{\nabla}_{Q} L)_i = \sum_{l = 1}^d (\nabla_{\text{YOSO}} L)_{i, l} \sum_{j = 1}^n \mathcal{B}(Q, K)_{i,j} (V_{j, l} \frac{\tau}{2} K_j)
\label{eq:backward-d2}
\end{equation}
Since it needs $d$ runs of a subroutine whose complexity is $O(m 2^{\tau} d)$ and $O(n m d)$ for memory and time respectively, its memory complexity is $O(m 2^{\tau} d^2)$, and time complexity is $O(n m d^2)$. The $d^2$ term in the memory complexity can be eliminated by repeatedly using the same hash tables $d^2$ times without increasing runtime, which improves the memory complexity to $O(m 2^{\tau})$. The overall complexity of our method relative to softmax self-attention is shown in Table \ref{tab:complexity}.
%
To address the quadratic dependence on $d$, 
we describe a scheme to estimate the quantity \eqref{eq:pseudo-gradient-q} with a cost that is linear in $d$ and similarly, for estimating \eqref{eq:gradient-q} in appendix. The models trained using the latter estimate are identified as *YOSO in our experiments.




\section{Experiments}

In this section, we analyze YOSO experimentally and evaluate its performance. In the previous section, we assumed that queries and keys are unit length and described how to make the strategy work. In the experiments, we found that simply applying a $\ell_2$ normalization on queries and keys and using $\tau$ as a hyperparameter does not degrade the performance and yet is more efficient to compute, so we use the simpler version in the experiments.

For empirical evaluations, we evaluate YOSO-Attention on BERT language model pretraining followed by GLUE downstream tasks finetuning. Then, we compare our method with other efficient Transformer baselines using a small version of BERT and the LRA benchmark. 
As a sanity check, we also include YOSO-Attention (YOSO-E) where the expected attention weights are directly computed using collision probability. This represents the behavior of YOSO when the number of hashes tends to infinite. We verified that in all tasks, 
YOSO-E behaves similarly as softmax self-attention. 
Further, we demonstrate that the performance of YOSO-m (YOSO-Attention using $m$ hashes) generally converges to YOSO-E as $m$ increases. Also, training using the backpropagation estimate in \eqref{eq:pseudo-gradient-q} (denoted YOSO) converges in all tasks, but the backpropagation estimate based on \eqref{eq:gradient-q} (denoted *YOSO) is slightly better. When compared to other efficient Transformer baselines, we show that our proposal performs favorably while maintaining high efficiency for both time and memory. Finally, we empirically verified that the approximation error of YOSO-m stays relatively flat as the sequence length increases. 

\subsection{Language Modeling}

\begin{table*}[h]
\begin{center}
\begin{small}
\begin{sc}
\begin{tabular}{lccccccc}
\toprule
Method & MLM & SOP & MRPC & SST-2 & QNLI & QQP & MNLI-m/mm  \\
\midrule
Softmax & 4.65 & 94.2 & 88.3 & 91.1  & 90.3 & 87.3 & 82.4/82.4 \\
\midrule
YOSO-E  & 4.54 & 94.4 & 88.1 & 92.3  & 90.1 & 87.3 & 82.2/82.9 \\
YOSO-64 & 4.79 & 94.2 & 88.1 & 91.5  & 89.5 & 87.0 & 81.6/81.6 \\
YOSO-32 & 4.89 & 93.5 & 87.3 & 90.9  & 89.0 & 86.3 & 80.5/80.7 \\
YOSO-16 & 5.14 & 92.8 & 87.1 & 90.7  & 88.3 & 85.3 & 79.6/79.5 \\
*YOSO-32 & 4.89 & 93.5 & 87.6 & 91.4 & 90.0 & 86.8 & 80.5/80.9 \\
*YOSO-16 & 5.02 & 93.4 & 87.7 & 90.8 & 88.9 & 86.7 & 80.6/80.5 \\
\bottomrule
\end{tabular}
\end{sc}
\end{small}
\end{center}
\caption{Dev set results on MLM and SOP pretraining and GLUE tasks for comparison to softmax self-attention in BERT-base setting. We report perplexity for MLM, F1 score for MRPC and QQP, and accuracy for others. }
\label{tab:glue}
\end{table*}

To evaluate YOSO, we follow the BERT language model pretraining procedure \citep{devlin2018bert} and evaluate the performance of our method on both intrinsic tasks and multiple downstream tasks in the GLUE benchmark. 

\begin{figure}
\centering
\vspace{-0.05in}
\includegraphics[height=1.88in]{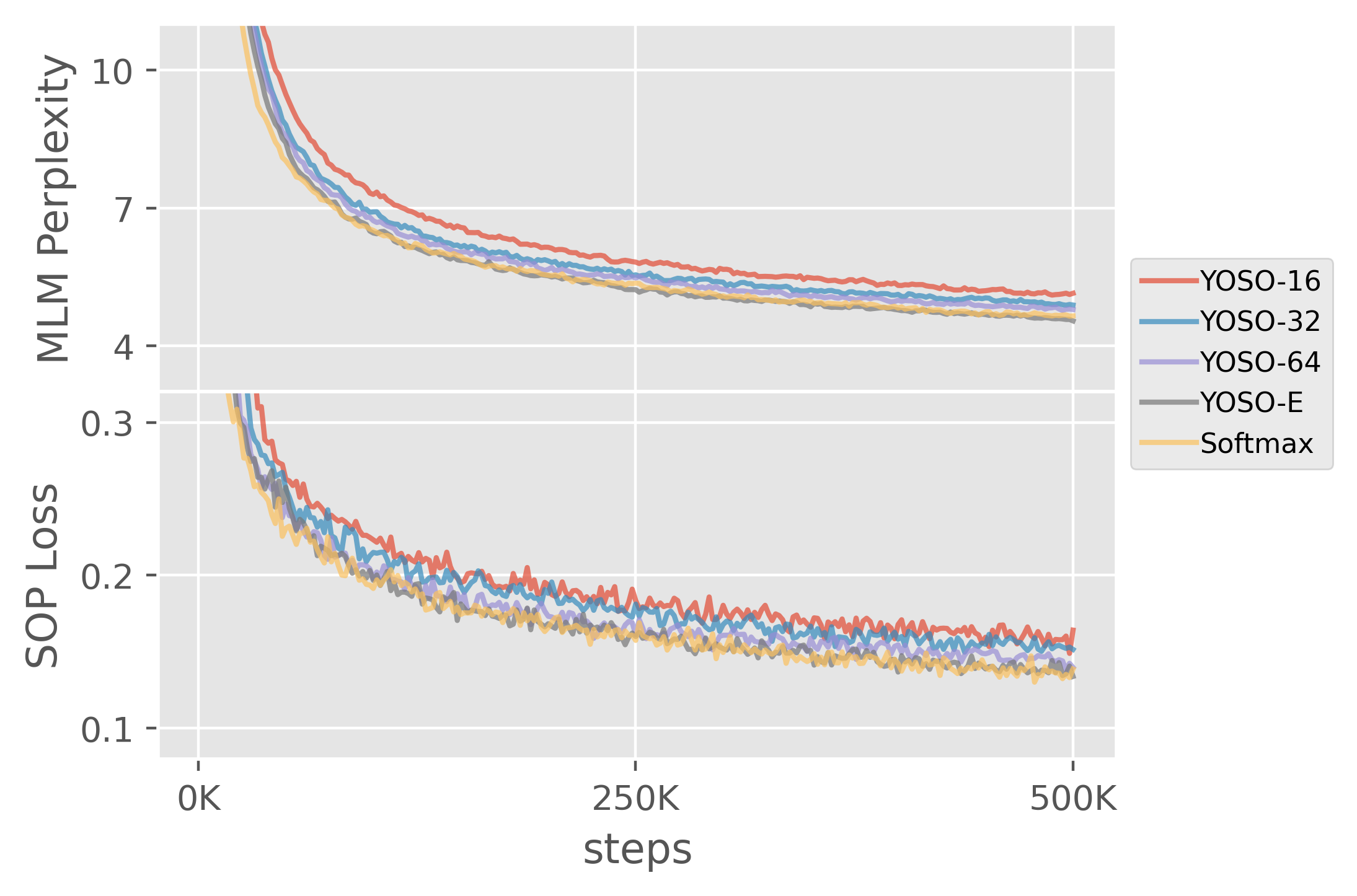}
\vspace{-12pt}
\caption{MLM and SOP result of $512$ sequence length language model pretraining. YOSO-x means the model is pretrained with YOSO-Attention using x hashes with E being expectation. }
\label{fig:pretrain-val}
\end{figure}

\textbf{BERT Pretraining}. Following \citet{devlin2018bert}, the model is pretrained on BookCorpus \citep{zhu2015aligning} and English Wikipedia. To evaluate the capacity of the model in capturing sentence level information, instead of using Next-Sentence-Prediction (NSP) as the sentence level loss as in the original BERT, we adapt the Sentence-Ordering-Prediction (SOP) from ALBERT \citep{lan2019albert} -- this is more difficult compared to NSP. All models are trained with Mask-Language-Modeling (MLM) and SOP objectives. We use the same hyperparameters for pretraining as \citet{devlin2018bert}. However, to keep the compute needs manageable, all models are trained for $500$K steps (batch size of $256$). 

\textbf{Number of Hashes during Pretraining}. 
Since the estimation variance decreases as the number of hashes increases, to evaluate the trade-off between efficiency and performance in YOSO, we test multiple hash settings: (*)YOSO-16, (*)YOSO-32, YOSO-64, and finally, YOSO-E (to simulate infinite hashes). We plot MLM validation perplexity and SOP validation loss curves of $512$ length models pretrained with softmax self-attention and YOSO-Attention (Fig. \ref{fig:pretrain-val} right) {and show the MLM validation perplexity and SOP accuracy obtained in Table \ref{tab:glue}}. The curves of YOSO-E agrees with and slightly exceeds softmax self-attention, indicating that YOSO is indeed as effective as self-attention. It is expected that as the number of hashes increase, the performance of YOSO will approach YOSO-E, as the approximation becomes more accurate. For both MLM and SOP, we confirm that YOSO is as effective as softmax self-attention. 

\begin{figure}[!b]
\centering
\vspace{-0.05in}
\includegraphics[height=1.88in]{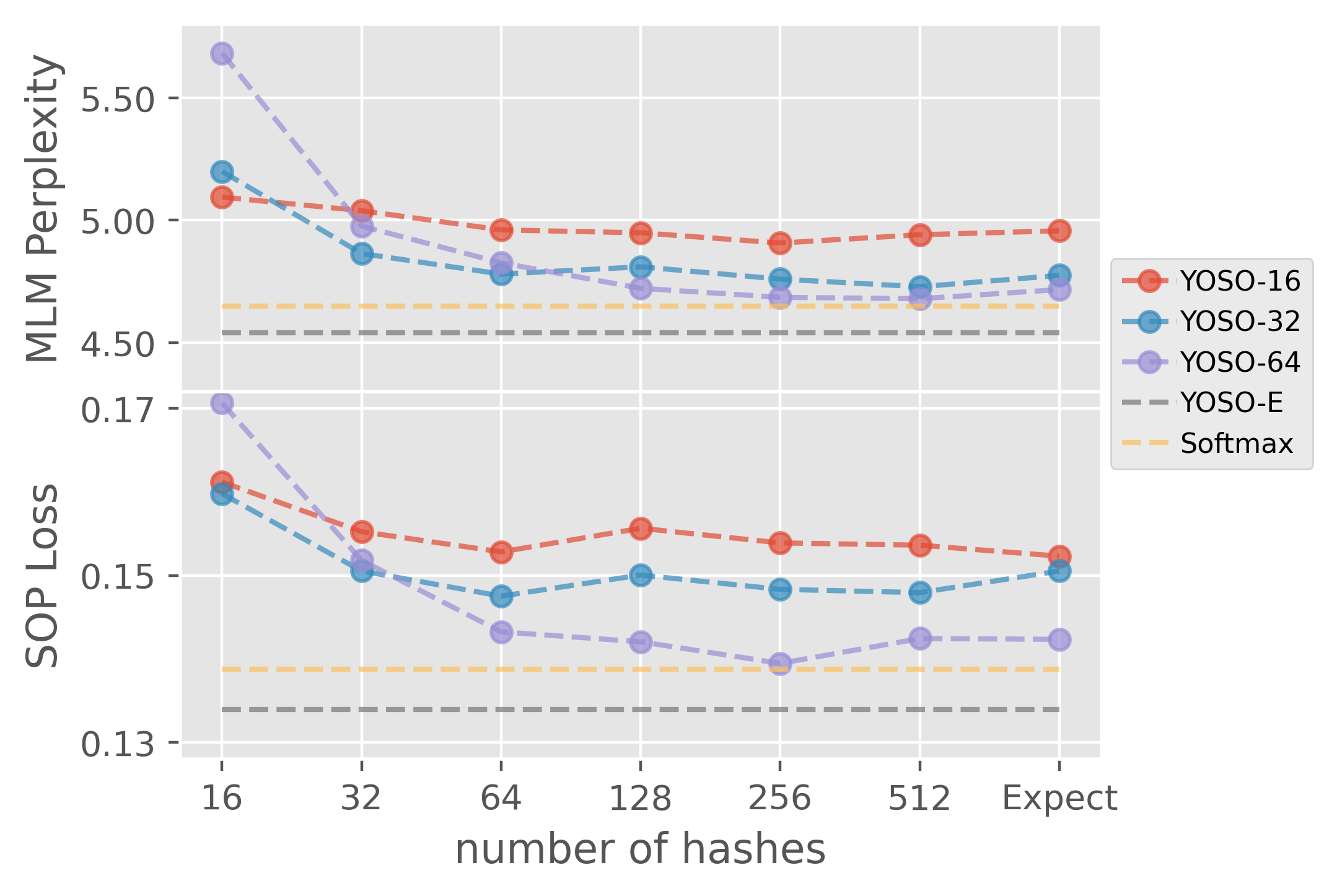}
\vspace{-12pt}
\caption{\label{fig:pretrained-val-diff-hash} MLM and SOP result of pretrained pretraining when altering the number of hashes in inference. }
\end{figure}

\textbf{Number of Hashes during Validation}. YOSO-Attention is a stochastic model. To make the inference deterministic, as in dropout \citep{srivastava2014dropout}, ideally, we take the expectation as the output. However, directly computing the expectation involves a $O(n^2)$ cost, so we experiment with the effect of different hash settings in validation and simulate expectation as the number of hashes increases. We plot the MLM perplexity and SOP loss of the same pretrained models using different number of hashes on validation in Figure \ref{fig:pretrained-val-diff-hash}. We observe that as the number of hashes increases, the MLM perplexity and SOP loss generally decreases for all pretraining hash settings.


\textbf{GLUE}. We examined the effectiveness of our method on diverse downstream tasks
and ask how YOSO compares with softmax self-attention even after finetuning. 
We finetuned all pretrained BERT-base model on 
MRPC \citep{dolan2005automatically}, SST-2 \citep{socher2013recursive}, 
QNLI \citep{rajpurkar2016squad}, QQP \citep{chen2018quora}, and MNLI \citep{williams2018broad} tasks in the GLUE benchmarks and report their corresponding dev metrics. For large datasets including QNLI, QQP, and MNLI, due to extensive resource needs, we did not perform hyperparameter search, so we used a batch size of 32 and learning rate 3e-5 to update our model and finetune our models for 4 epochs. For MRPC and SST-2, we follow BERT finetuning to do a hyperparameter search with candidate batch size $\{$8, 16, 32$\}$ and learning rate $\{$2e-5, 3e-5, 4e-5, 5e-5$\}$  and select the best dev set result. 
Results are listed in Table \ref{tab:glue}. We observed that YOSO's performance on downstream tasks is comparable with softmax self-attention, and even shows slightly better results in some hash settings. Further, the downstream performance of YOSO generally increases with more hashes, providing an adjustable trade-off between efficiency and accuracy.

\subsection{Performance considerations relative to baselines}

\begin{table*}[!htbp]
\begin{center}
\begin{small}
\begin{sc}
\begin{tabular}{lccccc|c}
\toprule
Method & listops & text & retrieval & image & pathfinder & Avg  \\
\tiny{Sequence Length} & \tiny{2K} & \tiny{4K}   & \tiny{4K}   & \tiny{1K}   & \tiny{1K} &   \\
\midrule
none & 19.20 & 61.11 & 74.78 & 33.86 & 66.51 & 51.09 \\
\midrule
softmax & 37.10 & 65.02 & 79.35 & 38.20 & 74.16 & 58.77 \\
yoso-e & 37.30 & 64.71 & 81.16 & 39.78 & 72.90 & 59.17 \\
\midrule
nystr\"{o}mformer & 37.15 & 65.52 & 79.56 & 41.58 & 70.94 & 58.95 \\
longformer & 37.20 & 64.60 & 80.97 & 39.06 & 73.01 & 58.97 \\
linformer & 37.25 & 55.91 & 79.37 & 37.84 & 67.60 & 55.59 \\
reformer & 19.05 & 64.88 & 78.64 & 43.29 & 69.36 & 55.04 \\
performer & 18.80 & 63.81 & 78.62 & 37.07 & 69.87 & 53.63 \\
yoso-32 & 37.25 & 63.12 & 78.69 & 40.21 & 72.33 & 58.32 \\
yoso-C-16 & 37.40 & 64.28 & 77.61 & 44.67 & 71.86 & 59.16 \\
*yoso-16 & 37.20 & 62.97 & 79.02 & 40.50 & 72.05 & 58.35 \\
*yoso-C-16 & 37.35 & 65.89 & 78.80 & 45.93 & 71.39 & 59.87 \\
\bottomrule
\end{tabular}
\end{sc}
\end{small}
\end{center}
\vspace{-12pt}
\caption{Test set accuracy of LRA tasks. 
Our proposed YOSO is comparable to Longformer and Nystr\"{o}mformer and outperforms other baselines, and YOSO with a depthwise convolution outperforms all baselines. 
}
\label{tab:lra}
\end{table*}

\begin{figure}[!thbp]
\centering
\vspace{-0.05in}
\includegraphics[height=1.88in]{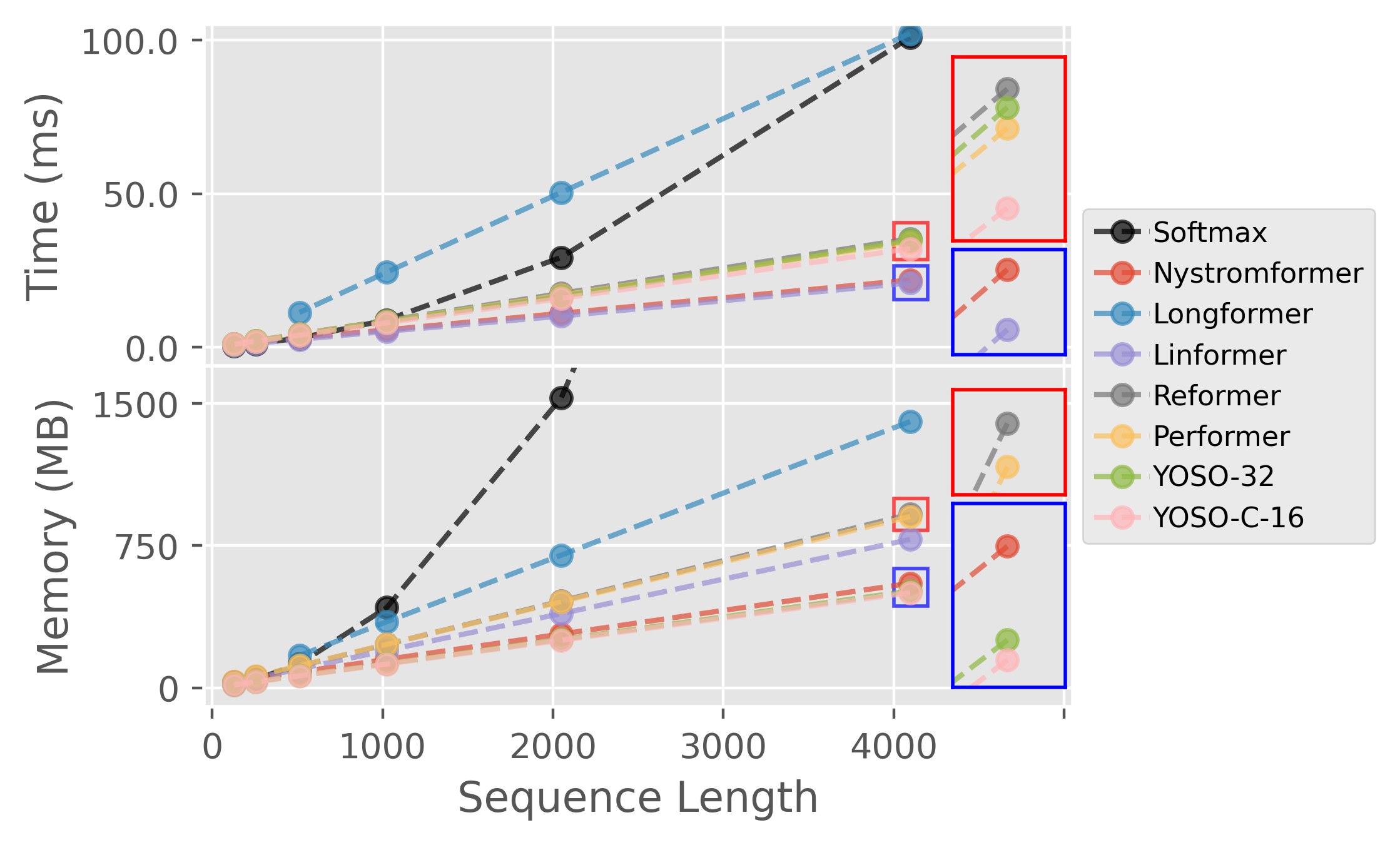}
\vspace{-12pt}
\caption{Running time and memory consumption results on various input sequence length. The reported values pertain to a single instance. Time is estimated by averaging total runtime and then dividing it by batch size, while memory is measured by dividing total memory consumption by batch size. Note that the trend lines are overlapping for several methods, see appendix for details. }
\label{fig:time-mem-comparison-max-batch}
\end{figure}

We also evaluate how well our method performs compared to other efficient Transformer baselines. For the baselines, we compared YOSO with Nystromformer ($64$ landmarks and $33$ convolution size) \citep{xiong2021nystromformer}, Longformer ($512$ attention window size) \citep{beltagy2020longformer}, Linformer ($256$ projection dimensions) \cite{Wang2020LinformerSW}, Reformer ($2$ hashes) \citep{Kitaev2020ReformerTE}, and Performer ($256$ random feature dimensions) \citep{choromanski2020rethinking} on a small version of BERT and LRA benchmark. The same model-specific hyperparameters are also used in efficiency profiles in Figure \ref{fig:time-mem-comparison-max-batch}. Further, inspired by Nystromformer, we also tested adding a depthwise convolution, which is referred as YOSO-C-x (x for the number of hashes). The experiment results indicate that depthwise convolution improves the performance of our method in some tasks.  

\textbf{BERT-Small}. For BERT pretraining task, due to the large resource needs of running all baselines in BERT-base setting, we evaluate all methods in a BERT-small setting (4 layers, 512 dimensions, 8 heads) with $500$K steps pretraining. Since the attention window size of Longformer is the same as the maximal sequence length of the input, it provides full self-attention, similar to softmax in this setting. Softmax self-attention achieves 7.05 MLM validation perplexity and 91.3\% SOP validation accuracy on this task, and YOSO (with convolution) achieves 7.34 MLM validation perplexity and 89.6\% SOP validation accuracy. Here, YOSO-C is comparable to softmax self-attention and Nystr\"{o}mformer while performs favorably relative to Reformer and Performer. 
For GLUE tasks, we found that 99\% of instances in MRPC, SST-2, QNLI, QQP, and MNLI have sequence lengths less than $112$. Since the chunked attention window in Reformer can capture full attention across all tokens in this setting, we expect to see similar performance for Reformer as softmax self-attention. We provide  results for QNLI, QQP, and MNLI for all baselines in the appendix (also includes results on MLM and SOP pretraining tasks).

\textbf{LRA Benchmark}. To evaluate the generalization of YOSO on diverse tasks and its viability on longer sequence tasks, we run our method on LRA benchmark \citep{tay2020long} and compare it with standard Transformer as well as other efficient Transformer baselines. This benchmark consists of five tasks: Listops \citep{nangia2018listops}, byte-level IMDb reviews classfication (Text) \citep{maas2011learning}, byte-level document matching (Retrieval) \citep{radev2013acl}, pixel-level CIFAR-10 classification (image) \citep{krizhevsky2009learning}, and pixel-level Pathfinder \citep{linsley2018learning}. These tasks are designed to assess different aspects of an efficient Transformer and provide a comprehensive analysis of its generalization on longer sequence tasks. 

Since the code release from \cite{tay2020long} only runs on TPUs, and the hyperparameters are not known, we followed the experimental settings in \citet{xiong2021nystromformer}. We include a model without self-attention, labeled ``None", as a reference to show how much each baseline helps in modeling long sequences. The results are shown in Table \ref{tab:lra}. The performance of YOSO compared to softmax self-attention on LRA tasks is consistent with the results we reported for language modeling. For baseline comparisons,  YOSO is comparable to Longformer and Nystr\"{o}mformer and outperforms all other baselines by 3\% average accuracy across the five tasks. Further, with a depthwise convolution, 
YOSO outperforms all baselines. These results provide direct evidence for the applicability of  YOSO on longer sequences.

\begin{figure*}[!htbp]
\centering
\vspace{-0.05in}
\includegraphics[width=0.975\textwidth]{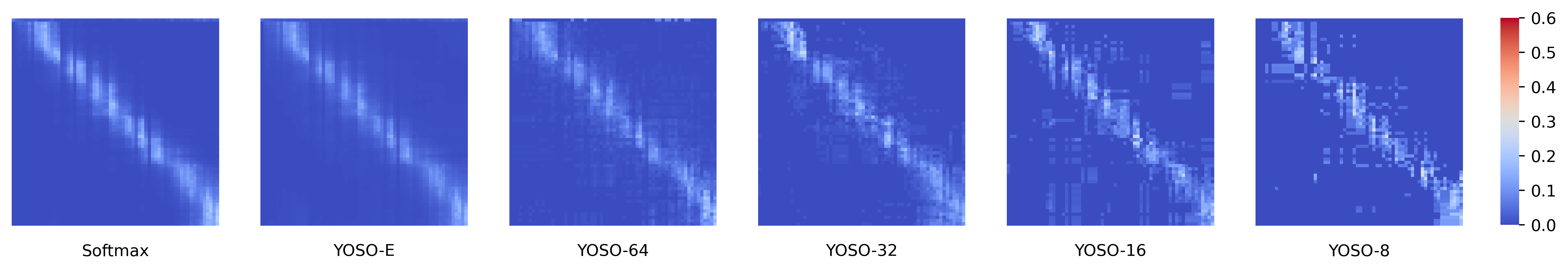}
\vspace{-12pt}
\caption{Attention matrices generated by Softmax and YOSO using the same input. We only visualize self-attention matrices for the first $64$ tokens. Notice that the patterns are preserved well.}
\label{fig:attn_maps}
\end{figure*}

\subsection{Efficiency considerations relative to baselines}


The overall thrust efficient Transformer models is to have the same capacity as a standard Transformer while reducing the time and memory cost of self-attention. In this section, we profile the running time and memory consumption of our method as well as vanilla Transformer and other efficient Transformers for different sequence lengths. We use a Transformer model of 6 layers, 256 embedding dimension, 1024 hidden dimension, 4 attention heads and measure runtime and peak memory consumption using random inputs. 
To achieve the best efficiency for each baseline, for each method and each sequence length, we use the largest batch size we can fit into memory and run training for 10 steps and average the results to estimate the time and memory cost of a single instance. The experiments were performed on a single NVIDIA 2080TI. The result is shown in Figure \ref{fig:time-mem-comparison-max-batch}. While Longformer scales linearly with respect to the sequence length, the benefit comes from longer sequence, which is consistent to \cite{beltagy2020longformer}. The detailed results and further experiments on efficiency are provided in the appendix. The profiling results indicate that our YOSO is able to scale efficiently with input sequence lengths. Further, the results suggest that our YOSO is highly efficient in terms of runtime and offers the highest efficiency in terms of memory compared to baselines.

\subsection{How large is the approximation error?}

\begin{figure}[!b]
\centering
\vspace{-0.05in}
\includegraphics[height=1.18in]{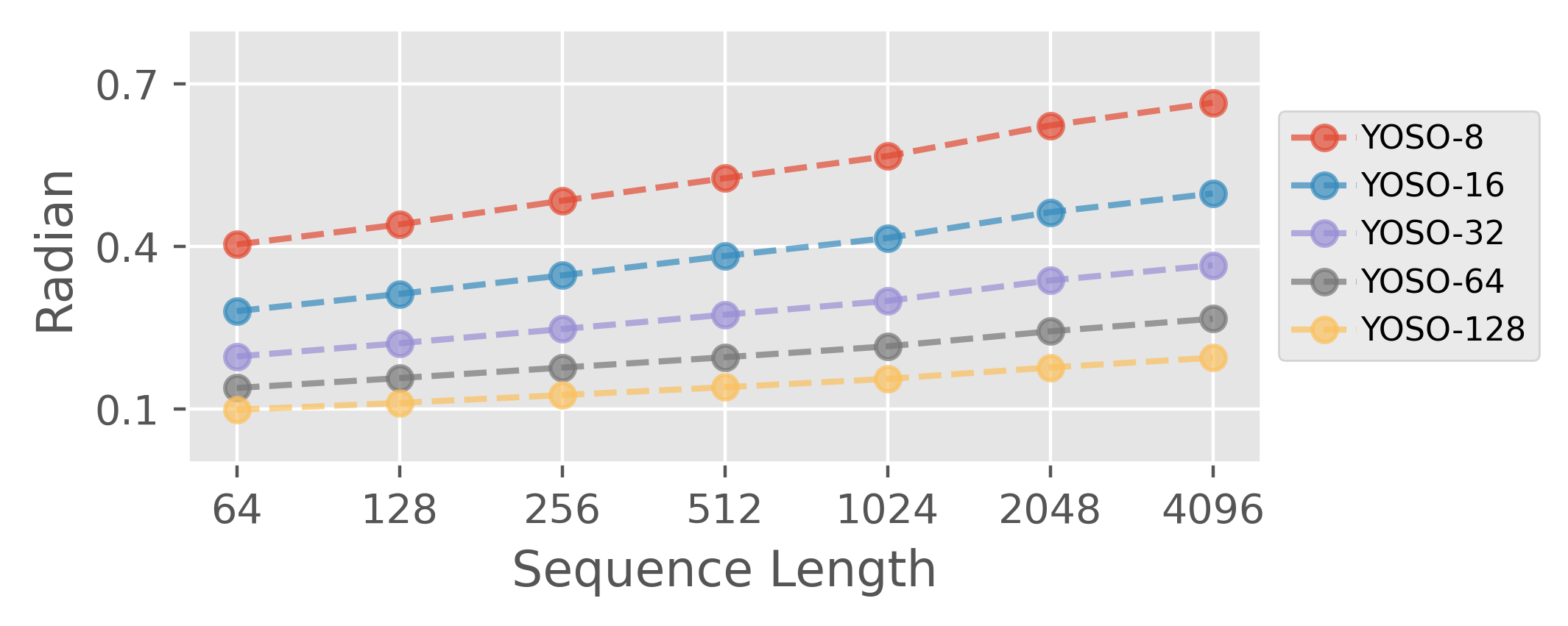}
\vspace{-12pt}
\caption{Averaged Radian between outputs of YOSO-E and YOSO-m for $m = 8, 16, 32, 64, 128$ for sequence length from $64$ to $4096$. The x axis is in logarithm scale, so we verify that the error only increase at a logarithm speed w.r.t. the sequence length.}
\label{fig:approx-error-diff-hash}
\end{figure}

To assess the estimation error of YOSO, we generate attention matrices of YOSO using $Q, K$ from a trained model and compare it against softmax self-attention. In Figure \ref{fig:attn_maps}, visually, our method produces similar attention patterns as softmax self-attention. The estimation of attention matrix is more accurate as the number of hashes increases. 
Further, in the formulation of YOSO, each output of YOSO-Attention is a weighted sum of random variables as shown in \eqref{eq:sum-rv}; so one may suspect that as the sequence length increases, the variance of YOSO-Attention output might potentially increase. To assess the increase in variance, we use $Q, K, V$ from a trained model and measure the averaged angle between YOSO-E and YOSO-m for $m = 8, 16, 32, 64, 128$ for sequence lengths between $64$ to $4096$. Since YOSO outputs unit vectors, only the vector direction is meaningful, we use the radian between outputs of YOSO-E and YOSO-m to assess the approximation error. The result is shown in Figure \ref{fig:approx-error-diff-hash}. The $x$-axis uses a logscale to verify that the approximation error increases at a much slower rate compared to the sequence length. One explanation is that most  attention weights are near zero, see Figure \ref{fig:attn_maps}. This means that the variance of the corresponding attention weights are near zero. In most cases, the size of the dependency (large attention weights) is relatively independent of the sequence length. As a result, the increase in sequence length does not introduce the same amount of error in approximation. 


\section{Conclusion}

We present a transformer-based model, YOSO-Attention, which scales linearly in the number of tokens. This 
allows YOSO to be applicable to long document tasks. 
Via 
a randomized sampling based scheme, YOSO approximates self-attention as a sum of individual tokens associated with Bernoulli random variables that can be sampled at once by a single hash, in principle. With specific modifications of LSH, YOSO-Attention can be efficiently deployed within a deep learning framework and various aspects of this idea and our implementation, we expect, will find use in 
some novel settings (e.g., point cloud modeling and vision).  
Our preliminary 
work suggests that YOSO has potential applications beyond Transformers, e.g., 
for scalability benefits in kernel density estimation and differentiable LSH.  

\section*{Acknowledgments}
This work was supported by the University of Wisconsin 
Data Science Institute through funding provided by American Family Insurance. 
YX and VS were also supported by 
University of Wisconsin Center for Predictive Computational Phenotyping (CPCP) 
funded by NIH U54 AI117924. SNR was supported by  UIC-ICR
start-up funds. The authors 
thank Rudrasis Chakraborty for many helpful discussions. 

\bibliography{main}

\begin{thebibliography}{35}
\providecommand{\natexlab}[1]{#1}
\providecommand{\url}[1]{\texttt{#1}}
\expandafter\ifx\csname urlstyle\endcsname\relax
  \providecommand{\doi}[1]{doi: #1}\else
  \providecommand{\doi}{doi: \begingroup \urlstyle{rm}\Url}\fi

\bibitem[Andoni et~al.(2015)Andoni, Indyk, Laarhoven, Razenshteyn, and
  Schmidt]{alex2015practical}
Andoni, A., Indyk, P., Laarhoven, T., Razenshteyn, I., and Schmidt, L.
\newblock Practical and optimal lsh for angular distance.
\newblock In Cortes, C., Lawrence, N., Lee, D., Sugiyama, M., and Garnett, R.
  (eds.), \emph{Advances in Neural Information Processing Systems}, volume~28.
  Curran Associates, Inc., 2015.
\newblock URL
  \url{https://proceedings.neurips.cc/paper/2015/file/2823f4797102ce1a1aec05359cc16dd9-Paper.pdf}.

\bibitem[Beltagy et~al.(2020)Beltagy, Peters, and Cohan]{beltagy2020longformer}
Beltagy, I., Peters, M.~E., and Cohan, A.
\newblock Longformer: The long-document transformer.
\newblock \emph{arXiv preprint arXiv:2004.05150}, 2020.

\bibitem[Charikar \& Siminelakis(2017)Charikar and
  Siminelakis]{charikar2017hashing}
Charikar, M. and Siminelakis, P.
\newblock Hashing-based-estimators for kernel density in high dimensions.
\newblock In \emph{2017 IEEE 58th Annual Symposium on Foundations of Computer
  Science (FOCS)}, pp.\  1032--1043, 2017.
\newblock \doi{10.1109/FOCS.2017.99}.

\bibitem[Charikar(2002)]{charikar2002similarity}
Charikar, M.~S.
\newblock Similarity estimation techniques from rounding algorithms.
\newblock In \emph{Proceedings of the Thiry-Fourth Annual ACM Symposium on
  Theory of Computing}, STOC '02, pp.\  380–388, New York, NY, USA, 2002.
  Association for Computing Machinery.
\newblock ISBN 1581134959.
\newblock \doi{10.1145/509907.509965}.
\newblock URL \url{https://doi.org/10.1145/509907.509965}.

\bibitem[Chen et~al.(2018)Chen, Zhang, Zhang, and Zhao]{chen2018quora}
Chen, Z., Zhang, H., Zhang, X., and Zhao, L.
\newblock Quora question pairs, 2018.

\bibitem[Child et~al.(2019)Child, Gray, Radford, and
  Sutskever]{Child2019GeneratingLS}
Child, R., Gray, S., Radford, A., and Sutskever, I.
\newblock Generating long sequences with sparse transformers.
\newblock \emph{arXiv preprint arXiv:1904.10509}, 2019.

\bibitem[Choromanski et~al.(2021)Choromanski, Likhosherstov, Dohan, Song, Gane,
  Sarlos, Hawkins, Davis, Mohiuddin, Kaiser, Belanger, Colwell, and
  Weller]{choromanski2020rethinking}
Choromanski, K.~M., Likhosherstov, V., Dohan, D., Song, X., Gane, A., Sarlos,
  T., Hawkins, P., Davis, J.~Q., Mohiuddin, A., Kaiser, L., Belanger, D.~B.,
  Colwell, L.~J., and Weller, A.
\newblock Rethinking attention with performers.
\newblock In \emph{International Conference on Learning Representations}, 2021.
\newblock URL \url{https://openreview.net/forum?id=Ua6zuk0WRH}.

\bibitem[Devlin et~al.(2019)Devlin, Chang, Lee, and Toutanova]{devlin2018bert}
Devlin, J., Chang, M.-W., Lee, K., and Toutanova, K.
\newblock {BERT}: Pre-training of deep bidirectional transformers for language
  understanding.
\newblock In \emph{Proceedings of the 2019 Conference of the North {A}merican
  Chapter of the Association for Computational Linguistics: Human Language
  Technologies, Volume 1 (Long and Short Papers)}, pp.\  4171--4186,
  Minneapolis, Minnesota, June 2019. Association for Computational Linguistics.
\newblock \doi{10.18653/v1/N19-1423}.
\newblock URL \url{https://www.aclweb.org/anthology/N19-1423}.

\bibitem[Dolan \& Brockett(2005)Dolan and Brockett]{dolan2005automatically}
Dolan, W.~B. and Brockett, C.
\newblock Automatically constructing a corpus of sentential paraphrases.
\newblock In \emph{Proceedings of the Third International Workshop on
  Paraphrasing ({IWP}2005)}, 2005.
\newblock URL \url{https://www.aclweb.org/anthology/I05-5002}.

\bibitem[Katharopoulos et~al.(2020)Katharopoulos, Vyas, Pappas, and
  Fleuret]{Katharopoulos2020TransformersAR}
Katharopoulos, A., Vyas, A., Pappas, N., and Fleuret, F.
\newblock Transformers are {RNN}s: Fast autoregressive transformers with linear
  attention.
\newblock In III, H.~D. and Singh, A. (eds.), \emph{Proceedings of the 37th
  International Conference on Machine Learning}, volume 119 of
  \emph{Proceedings of Machine Learning Research}, pp.\  5156--5165. PMLR,
  13--18 Jul 2020.
\newblock URL \url{http://proceedings.mlr.press/v119/katharopoulos20a.html}.

\bibitem[Kitaev et~al.(2020)Kitaev, Kaiser, and Levskaya]{Kitaev2020ReformerTE}
Kitaev, N., Kaiser, L., and Levskaya, A.
\newblock Reformer: The efficient transformer.
\newblock In \emph{International Conference on Learning Representations}, 2020.
\newblock URL \url{https://openreview.net/forum?id=rkgNKkHtvB}.

\bibitem[Krizhevsky et~al.(2009)Krizhevsky, Hinton,
  et~al.]{krizhevsky2009learning}
Krizhevsky, A., Hinton, G., et~al.
\newblock Learning multiple layers of features from tiny images.
\newblock 2009.

\bibitem[Lan et~al.(2020)Lan, Chen, Goodman, Gimpel, Sharma, and
  Soricut]{lan2019albert}
Lan, Z., Chen, M., Goodman, S., Gimpel, K., Sharma, P., and Soricut, R.
\newblock Albert: A lite bert for self-supervised learning of language
  representations.
\newblock In \emph{International Conference on Learning Representations}, 2020.
\newblock URL \url{https://openreview.net/forum?id=H1eA7AEtvS}.

\bibitem[Levy et~al.(2015)Levy, Goldberg, and Dagan]{levy-etal-2015-improving}
Levy, O., Goldberg, Y., and Dagan, I.
\newblock Improving distributional similarity with lessons learned from word
  embeddings.
\newblock \emph{Transactions of the Association for Computational Linguistics},
  3:\penalty0 211--225, 2015.
\newblock \doi{10.1162/tacl_a_00134}.
\newblock URL \url{https://www.aclweb.org/anthology/Q15-1016}.

\bibitem[Linsley et~al.(2018)Linsley, Kim, Veerabadran, Windolf, and
  Serre]{linsley2018learning}
Linsley, D., Kim, J., Veerabadran, V., Windolf, C., and Serre, T.
\newblock Learning long-range spatial dependencies with horizontal gated
  recurrent units.
\newblock In Bengio, S., Wallach, H., Larochelle, H., Grauman, K.,
  Cesa-Bianchi, N., and Garnett, R. (eds.), \emph{Advances in Neural
  Information Processing Systems}, volume~31. Curran Associates, Inc., 2018.
\newblock URL
  \url{https://proceedings.neurips.cc/paper/2018/file/ec8956637a99787bd197eacd77acce5e-Paper.pdf}.

\bibitem[Liu et~al.(2019)Liu, Ott, Goyal, Du, Joshi, Chen, Levy, Lewis,
  Zettlemoyer, and Stoyanov]{liu2019roberta}
Liu, Y., Ott, M., Goyal, N., Du, J., Joshi, M., Chen, D., Levy, O., Lewis, M.,
  Zettlemoyer, L., and Stoyanov, V.
\newblock Roberta: A robustly optimized bert pretraining approach.
\newblock \emph{arXiv preprint arXiv:1907.11692}, 2019.

\bibitem[Maas et~al.(2011)Maas, Daly, Pham, Huang, Ng, and
  Potts]{maas2011learning}
Maas, A.~L., Daly, R.~E., Pham, P.~T., Huang, D., Ng, A.~Y., and Potts, C.
\newblock Learning word vectors for sentiment analysis.
\newblock In \emph{Proceedings of the 49th Annual Meeting of the Association
  for Computational Linguistics: Human Language Technologies}, pp.\  142--150,
  Portland, Oregon, USA, June 2011. Association for Computational Linguistics.
\newblock URL \url{https://www.aclweb.org/anthology/P11-1015}.

\bibitem[Nangia \& Bowman(2018)Nangia and Bowman]{nangia2018listops}
Nangia, N. and Bowman, S.
\newblock {L}ist{O}ps: A diagnostic dataset for latent tree learning.
\newblock In \emph{Proceedings of the 2018 Conference of the North {A}merican
  Chapter of the Association for Computational Linguistics: Student Research
  Workshop}, pp.\  92--99, New Orleans, Louisiana, USA, June 2018. Association
  for Computational Linguistics.
\newblock \doi{10.18653/v1/N18-4013}.
\newblock URL \url{https://www.aclweb.org/anthology/N18-4013}.

\bibitem[Neyshabur \& Srebro(2015)Neyshabur and Srebro]{neyshabur2015symmetric}
Neyshabur, B. and Srebro, N.
\newblock On symmetric and asymmetric lshs for inner product search.
\newblock In Bach, F. and Blei, D. (eds.), \emph{Proceedings of the 32nd
  International Conference on Machine Learning}, volume~37 of \emph{Proceedings
  of Machine Learning Research}, pp.\  1926--1934, Lille, France, 07--09 Jul
  2015. PMLR.
\newblock URL \url{http://proceedings.mlr.press/v37/neyshabur15.html}.

\bibitem[Peng et~al.(2021)Peng, Pappas, Yogatama, Schwartz, Smith, and
  Kong]{peng2021rfa}
Peng, H., Pappas, N., Yogatama, D., Schwartz, R., Smith, N., and Kong, L.
\newblock Random feature attention.
\newblock In \emph{International Conference on Learning Representations}, 2021.
\newblock URL \url{https://openreview.net/forum?id=QtTKTdVrFBB}.

\bibitem[Press et~al.(2007)Press, Teukolsky, Vetterling, and
  Flannery]{press2007numerical}
Press, W.~H., Teukolsky, S.~A., Vetterling, W.~T., and Flannery, B.~P.
\newblock \emph{Numerical recipes: the art of scientific computing, 3rd
  Edition}.
\newblock Cambridge University Press, 2007.
\newblock ISBN 9780521706858.

\bibitem[Radev et~al.(2013)Radev, Muthukrishnan, Qazvinian, and
  Abu{-}Jbara]{radev2013acl}
Radev, D.~R., Muthukrishnan, P., Qazvinian, V., and Abu{-}Jbara, A.
\newblock The {ACL} anthology network corpus.
\newblock \emph{Language Resources and Evaluation}, 47\penalty0 (4):\penalty0
  919--944, 2013.
\newblock \doi{10.1007/s10579-012-9211-2}.

\bibitem[Raffel et~al.(2020)Raffel, Shazeer, Roberts, Lee, Narang, Matena,
  Zhou, Li, and Liu]{raffel2020exploring}
Raffel, C., Shazeer, N., Roberts, A., Lee, K., Narang, S., Matena, M., Zhou,
  Y., Li, W., and Liu, P.~J.
\newblock Exploring the limits of transfer learning with a unified text-to-text
  transformer.
\newblock \emph{Journal of Machine Learning Research}, 21\penalty0
  (140):\penalty0 1--67, 2020.
\newblock URL \url{http://jmlr.org/papers/v21/20-074.html}.

\bibitem[Rajpurkar et~al.(2016)Rajpurkar, Zhang, Lopyrev, and
  Liang]{rajpurkar2016squad}
Rajpurkar, P., Zhang, J., Lopyrev, K., and Liang, P.
\newblock {SQ}u{AD}: 100,000+ questions for machine comprehension of text.
\newblock In \emph{Proceedings of the 2016 Conference on Empirical Methods in
  Natural Language Processing}, pp.\  2383--2392, Austin, Texas, November 2016.
  Association for Computational Linguistics.
\newblock \doi{10.18653/v1/D16-1264}.
\newblock URL \url{https://www.aclweb.org/anthology/D16-1264}.

\bibitem[Socher et~al.(2013)Socher, Perelygin, Wu, Chuang, Manning, Ng, and
  Potts]{socher2013recursive}
Socher, R., Perelygin, A., Wu, J., Chuang, J., Manning, C.~D., Ng, A., and
  Potts, C.
\newblock Recursive deep models for semantic compositionality over a sentiment
  treebank.
\newblock In \emph{Proceedings of the 2013 Conference on Empirical Methods in
  Natural Language Processing}, pp.\  1631--1642, Seattle, Washington, USA,
  October 2013. Association for Computational Linguistics.
\newblock URL \url{https://www.aclweb.org/anthology/D13-1170}.

\bibitem[Spring \& Shrivastava(2018)Spring and Shrivastava]{spring2017new}
Spring, R. and Shrivastava, A.
\newblock Scalable estimation via {LSH} samplers {(LSS)}.
\newblock In \emph{International Conference on Learning Representations,
  Workshop Track Proceedings}, 2018.
\newblock URL \url{https://openreview.net/forum?id=BJazbHkPG}.

\bibitem[Srivastava et~al.(2014)Srivastava, Hinton, Krizhevsky, Sutskever, and
  Salakhutdinov]{srivastava2014dropout}
Srivastava, N., Hinton, G., Krizhevsky, A., Sutskever, I., and Salakhutdinov,
  R.
\newblock Dropout: A simple way to prevent neural networks from overfitting.
\newblock \emph{Journal of Machine Learning Research}, 15\penalty0
  (56):\penalty0 1929--1958, 2014.
\newblock URL \url{http://jmlr.org/papers/v15/srivastava14a.html}.

\bibitem[Tay et~al.(2021)Tay, Dehghani, Abnar, Shen, Bahri, Pham, Rao, Yang,
  Ruder, and Metzler]{tay2020long}
Tay, Y., Dehghani, M., Abnar, S., Shen, Y., Bahri, D., Pham, P., Rao, J., Yang,
  L., Ruder, S., and Metzler, D.
\newblock Long range arena : A benchmark for efficient transformers.
\newblock In \emph{International Conference on Learning Representations}, 2021.
\newblock URL \url{https://openreview.net/forum?id=qVyeW-grC2k}.

\bibitem[Vaswani et~al.(2017)Vaswani, Shazeer, Parmar, Uszkoreit, Jones, Gomez,
  Kaiser, and Polosukhin]{vaswani2017attention}
Vaswani, A., Shazeer, N., Parmar, N., Uszkoreit, J., Jones, L., Gomez, A.~N.,
  Kaiser, L.~u., and Polosukhin, I.
\newblock Attention is all you need.
\newblock In Guyon, I., Luxburg, U.~V., Bengio, S., Wallach, H., Fergus, R.,
  Vishwanathan, S., and Garnett, R. (eds.), \emph{Advances in Neural
  Information Processing Systems}, volume~30. Curran Associates, Inc., 2017.
\newblock URL
  \url{https://proceedings.neurips.cc/paper/2017/file/3f5ee243547dee91fbd053c1c4a845aa-Paper.pdf}.

\bibitem[Wang et~al.(2020)Wang, Li, Khabsa, Fang, and Ma]{Wang2020LinformerSW}
Wang, S., Li, B., Khabsa, M., Fang, H., and Ma, H.
\newblock Linformer: Self-attention with linear complexity.
\newblock \emph{arXiv preprint arXiv:2006.04768}, 2020.

\bibitem[Williams et~al.(2018)Williams, Nangia, and Bowman]{williams2018broad}
Williams, A., Nangia, N., and Bowman, S.
\newblock A broad-coverage challenge corpus for sentence understanding through
  inference.
\newblock In \emph{Proceedings of the 2018 Conference of the North {A}merican
  Chapter of the Association for Computational Linguistics: Human Language
  Technologies, Volume 1 (Long Papers)}, pp.\  1112--1122, New Orleans,
  Louisiana, June 2018. Association for Computational Linguistics.
\newblock \doi{10.18653/v1/N18-1101}.
\newblock URL \url{https://www.aclweb.org/anthology/N18-1101}.

\bibitem[Xiong et~al.(2021)Xiong, Zeng, Chakraborty, Tan, Fung, Li, and
  Singh]{xiong2021nystromformer}
Xiong, Y., Zeng, Z., Chakraborty, R., Tan, M., Fung, G., Li, Y., and Singh, V.
\newblock Nyströmformer: A nyström-based algorithm for approximating
  self-attention.
\newblock \emph{Proceedings of the AAAI Conference on Artificial Intelligence},
  35\penalty0 (16):\penalty0 14138--14148, May 2021.
\newblock URL \url{https://ojs.aaai.org/index.php/AAAI/article/view/17664}.

\bibitem[Yang et~al.(2019)Yang, Dai, Yang, Carbonell, Salakhutdinov, and
  Le]{yang2020xlnet}
Yang, Z., Dai, Z., Yang, Y., Carbonell, J., Salakhutdinov, R.~R., and Le, Q.~V.
\newblock Xlnet: Generalized autoregressive pretraining for language
  understanding.
\newblock In Wallach, H., Larochelle, H., Beygelzimer, A., d\textquotesingle
  Alch\'{e}-Buc, F., Fox, E., and Garnett, R. (eds.), \emph{Advances in Neural
  Information Processing Systems}, volume~32. Curran Associates, Inc., 2019.
\newblock URL
  \url{https://proceedings.neurips.cc/paper/2019/file/dc6a7e655d7e5840e66733e9ee67cc69-Paper.pdf}.

\bibitem[Zaheer et~al.(2020)Zaheer, Guruganesh, Dubey, Ainslie, Alberti,
  Ontanon, Pham, Ravula, Wang, Yang, and Ahmed]{zaheer2020big}
Zaheer, M., Guruganesh, G., Dubey, K.~A., Ainslie, J., Alberti, C., Ontanon,
  S., Pham, P., Ravula, A., Wang, Q., Yang, L., and Ahmed, A.
\newblock Big bird: Transformers for longer sequences.
\newblock In Larochelle, H., Ranzato, M., Hadsell, R., Balcan, M.~F., and Lin,
  H. (eds.), \emph{Advances in Neural Information Processing Systems},
  volume~33, pp.\  17283--17297. Curran Associates, Inc., 2020.
\newblock URL
  \url{https://proceedings.neurips.cc/paper/2020/file/c8512d142a2d849725f31a9a7a361ab9-Paper.pdf}.

\bibitem[Zhu et~al.(2015)Zhu, Kiros, Zemel, Salakhutdinov, Urtasun, Torralba,
  and Fidler]{zhu2015aligning}
Zhu, Y., Kiros, R., Zemel, R., Salakhutdinov, R., Urtasun, R., Torralba, A.,
  and Fidler, S.
\newblock Aligning books and movies: Towards story-like visual explanations by
  watching movies and reading books.
\newblock In \emph{2015 IEEE International Conference on Computer Vision
  (ICCV)}, pp.\  19--27, 2015.
\newblock \doi{10.1109/ICCV.2015.11}.

\end{thebibliography}
\bibliographystyle{icml2021}

\end{document}